\documentclass[conference]{IEEEtran}
\IEEEoverridecommandlockouts

\usepackage{fancyhdr}

\fancypagestyle{firstpage}{%
  \fancyhead{}
  
  \fancyfoot[C]{Accepted for Proceedings of IJCNN 2025, Rome, Italy. Copyright IEEE}
}
\usepackage{cite}
\usepackage{float}
\usepackage{amsmath,amssymb,amsfonts}
\usepackage[ruled,vlined]{algorithm2e}
\usepackage{graphicx}
\usepackage{gensymb}
\usepackage{textcomp}
\usepackage{xcolor}
\usepackage{steinmetz}
\usepackage{makecell}
\usepackage{tabularx}
\usepackage{balance}
\usepackage{subcaption}
\def\BibTeX{{\rm B\kern-.05em{\sc i\kern-.025em b}\kern-.08em
    T\kern-.1667em\lower.7ex\hbox{E}\kern-.125emX}}
\begin{document}

\title{Improved Accuracy of Robot Localization Using 3-D LiDAR in a Hippocampus-Inspired Model}

\author{\IEEEauthorblockN{Andrew Gerstenslager}
\IEEEauthorblockA{\textit{Dept. of Computer Science}\\
\textit{University of Cincinnati}\\
Cincinnati, USA \\
gerstead@mail.uc.edu}
\and
\IEEEauthorblockN{Bekarys Dukenbaev}
\IEEEauthorblockA{\textit{Dept. of Computer Science}\\
\textit{University of Cincinnati}\\
Cincinnati, USA \\
dukenbbs@mail.uc.edu}
\and
\IEEEauthorblockN{Ali A. Minai}
\IEEEauthorblockA{\textit{Dept. of Electrical} \\
\textit{and Computer Engineering}\\
\textit{University of Cincinnati}\\
Cincinnati, USA \\
minaiaa@ucmail.uc.edu}
}

\maketitle

\thispagestyle{firstpage}


\begin{abstract}
Boundary Vector Cells (BVCs) are a class of neurons in the brains of vertebrates that encode environmental boundaries at specific distances and allocentric directions, playing a central role in forming place fields in the hippocampus. Most computational BVC models are restricted to two-dimensional (2D) environments, making them prone to spatial ambiguities in the presence of horizontal symmetries in the environment. To address this limitation, we incorporate vertical angular sensitivity into the BVC framework, thereby enabling robust boundary detection in three dimensions, and leading to significantly more accurate spatial localization in a biologically-inspired robot model.

The proposed model processes LiDAR data to capture vertical contours, thereby disambiguating locations that would be indistinguishable under a purely 2D representation. Experimental results show that in environments with minimal vertical variation, the proposed 3D model matches the performance of a 2D baseline; yet, as 3D complexity increases, it yields substantially more distinct place fields and markedly reduces spatial aliasing. These findings show that adding a vertical dimension to BVC-based localization can significantly enhance navigation and mapping in real-world 3D spaces while retaining performance parity in simpler, near-planar scenarios.
\end{abstract}


\section{Introduction}
The hippocampus has been studied extensively for its role in enabling mammals to represent, localize, and navigate in new and familiar environments based on location-sensitive {\it Place Cells}. Specialized neurons such as {\it Head Direction Cells} (HDCs) \cite{taube1990head,taube1998head,taube2007head}, {\it Boundary Vector Cells} (BVCs) \cite{barry2006boundary,lever2009boundary,grieves2018boundary}, and {\it grid cells} \cite{hafting2005microstructure,solstad2006grid,Fyhn2007remapping,moser2008place,bush2014grid} in neighboring regions provide critical spatial information to {\it place cells} in the hippocampus proper \cite{Okeefe1978,OKeefe1987,McNaughton1987,muller1987spatial,redish1997cognitive,best2001annrevhippo}, which form a foundational framework for localization tasks. BVCs, in particular, are tuned to detect boundaries at specific distances and allocentric directions, while place cells encode specific locations based on input from BVCs \cite{barry2006boundary,solstad2006grid} and other spatially-tuned cells. Allocentric direction refers to angles defined relative to a global reference frame (e.g., north or fixed environmental landmarks), independent of the agent’s current heading. This contrasts with egocentric direction, which depends on the agent’s orientation. Together, these systems enable animals -- especially mammals -- to create internal maps of their environments.

Despite significant advances in understanding these mechanisms, most computational models of BVCs and place cells assume a 2D environment, ignoring the inherent three-dimensional complexity of real-world spaces. Such models often fail to distinguish locations that share similar 2D boundaries but differ vertically — an issue especially pronounced in robotics tasks like aerial navigation or multi-level indoor mapping. In complex 3D settings, leveraging vertical information is critical for precise localization and path planning.

To address this limitation, we extend the classical BVC model by introducing vertical sensitivity. The proposed model incorporates an additional Gaussian tuning parameter for vertical angular sensitivity, allowing the BVCs to process not just horizontal but also vertical information about boundaries. The firing rate of a BVC neuron is now determined by the distance, horizontal angle, and vertical angle relative to environmental boundaries. This enhancement enables the model to disambiguate locations with overlapping features in 2D by leveraging the added 3D spatial information. To our knowledge, this is the first model to directly extend the field-based BVC equations into 3D using the product of Gaussians over distance, azimuth, and elevation.

We implement the model specifically in the context of a robot that navigates along a 2D plane and uses 3D LiDAR for detecting boundaries and obstacles, and show that it successfully exploits 3D information in the environment improve localization based on place fields.


\section{Background}

\subsection{Biological Basis of Spatial Cognition}
The hippocampus and its interconnected neural circuits are central to spatial navigation and memory, with specialized neurons providing critical information for localization. 
place cells, first discovered by O'Keefe and Dostrovsky \cite{o1971hippocampus}, encode specific locations within an environment, forming a neural map of the space through their discrete firing fields. These fields have been studied extensively in 2D environments, showing that many neurons in the CA1 and CA3 regions of the hippocampus fire within distinct convex regions in each specific environment.  \cite{Okeefe1978,OKeefe1987,McNaughton1987,muller1987spatial,redish1997cognitive,best2001annrevhippo}.

BVCs complement these systems by responding to environmental boundaries at specific distances and allocentric directions, directly influencing place cell activations. 
The interaction between these systems underpins navigation and localization in mammals, as evidenced by behavioral and physiological studies\cite{barry2006boundary,lever2009boundary,grieves2018boundary}.

\subsection{3D Spatial Encoding in Animals}

While most research has focused on 2D spatial encoding, real-world navigation demands 3D spatial representation, a topic that has gained increasing attention recently. Among terrestrial species, rats have been extensively studied, providing evidence for the existence of 3D place fields. 
Research by Grieves et al. \cite{grieves2019,Grieves2020,grieves2021} demonstrated that place cells in rats exploring a cubic lattice maze form 3D firing fields that encode the entire volumetric space, albeit with anisotropic characteristics — with vertical encoding less stable compared to horizontal encoding due to the difficulty of vertical locomotion\cite{grieves2019, jovalekic2011}. 
Other studies have also observed that rat place fields on vertical surfaces tended to be elongated along the vertical axis, further emphasizing the challenges of 3D spatial representation in surface-traveling animals \cite{hayman2011,jeffery2013,jeffery2015}.

In aquatic environments, goldfish provide a compelling model for 3D spatial encoding due to their need to navigate freely in a volumetric medium. Although goldfish do not have a hippocampus, Cohen et al showed that they do contain boundary-sensitive neurons. These neurons are sensitive to specific distances and at both horizontal and vertical directions, allowing for a complete 3D encoding of environments\cite{cohen2023}. This capability underscores the critical role of 3D boundary detection in species that are not constrained to planar movement.

Bats, as flying mammals, exemplify volumetric navigation, with place cells forming isotropic 3D fields. Yartsev and Ulanovsky found that these fields were spherical, indicating equal sensitivity to all three dimensions\cite{Yartsev2013}.
Additionally, the head-direction cells in bats can encode both azimuth and pitch, supporting a fully 3D compass system\cite{jeffery2013,jeffery2015}.

These findings collectively highlight species-specific adaptations in 3D spatial encoding, reflecting ecological and behavioral demands. In rats, anisotropic encoding aligns with their terrestrial lifestyle, while goldfish and bats demonstrate isotropic encoding suitable for their volumetric environments.

\vspace{-2mm}
\subsection{Place Cell Aliasing}
It is well established that place cells use boundary information from BVCs to activate specific fields. Studies also show that place cells exhibit a high degree of spatial repetition, creating multiple fields in geometrically and visually similar regions\cite{spiers2015place, grieves2018}. This is known as {\it spatial aliasing}.

\begin{figure}[H]
    \centering
    \begin{subfigure}[b]{0.38\linewidth}
        \centering
        \includegraphics[width=\linewidth]{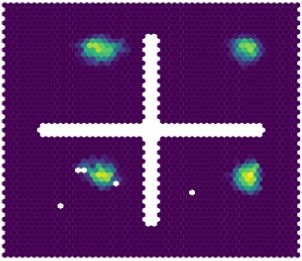}
        \caption{Multimodal place cell}
        \label{fig:multimodal_cell}
    \end{subfigure} \quad
    \begin{subfigure}[b]{0.38\linewidth}
        \centering
        \includegraphics[width=\linewidth]{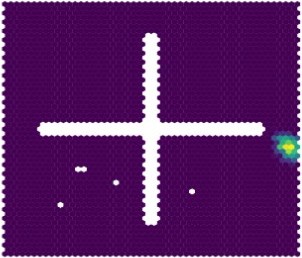}
        \caption{Unimodal place cell}
        \label{fig:unimodal_cell}
    \end{subfigure}
    
    \vspace{-1mm}
    \caption{Firing patterns of two place cells in an environment with cross-shaped boundaries.~\ref{fig:multimodal_cell} shows the firing pattern of an aliased place cell with four distinct areas of activation.~\ref{fig:unimodal_cell} shows the firing pattern of a unimodal place cell.}
    \label{fig:place_field_comparison}
\end{figure}

 Consistent with these findings, aliasing has been observed in continued work on the computational model developed by Alabi et. al \cite{alabi2020one, alabi2023rapid} in environments containing obstacles and distinct rooms where BVC activations can be found to be similar in multiple locations. To highlight this aliasing behavior, Figure~\ref{fig:place_field_comparison} contrasts two place cell responses: 
 (1) a multimodal place cell activation resulting from geometrically similar regions of the environment, and (2) a unimodal activation formed in a geometrically distinct region of the environment.

\vspace{-3mm}
\subsection{Motivation for Vertical Sensitivity}
The need for 3D spatial encoding is evident in the advancement of hippocampal-based computational models. This need is especially relevant in environments with complex vertical structures or with agents with movement capabilities beyond planar traversal.
In the following section, we will discuss the original computational model as well as the proposed improvements to add a tuning parameter for vertical sensitivity.


\vspace{-0.5mm}
\section{The Model}

\begin{figure}[H]
    \centering
    \includegraphics[width=70mm]{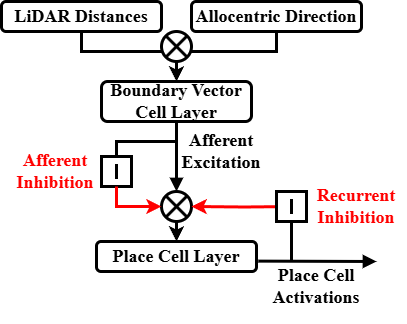}
    \vspace{-4mm}
    \caption{System Architecture. The BVC layer processes LiDAR boundary distances and agent orientation, supplying excitatory and inhibitory input to the PCN. The PCN applies recurrent inhibition to maintain a well-distributed place representation.}
    \label{fig:BVC_and_PCN}
\end{figure}

The architecture of the model is shown in Figure~\ref{fig:BVC_and_PCN}. It builds on one developed previously by our research group \cite{alabi2020one, alabi2023rapid}. The place cell forming component of the model comprises two computational layers: the BVC layer and the {\it place cell network} (PCN). The BVC layer encodes information about environmental boundaries by responding to boundary points at specific distances and angles from environmental boundary data, captured through LiDAR scans. The PCN integrates input from the BVCs and from recurrent connections between place cells to generate place fields for locations within the environment. Since BVCs are tuned to a fixed allocentric directions \cite{barry2006boundary}, the LiDAR data is preprocessed to align each scan with a global orientation. This ensures that environmental boundaries are consistently represented in an allocentric frame of reference, independent of the agent's orientation.

\vspace{-2mm}

\subsection{BVC Layer}

The BVC model used is an extension of one originally proposed by Barry et al. \cite{barry2006boundary}, and has neurons that exhibit Gaussian-tuned responses, with maximal activation when an obstacle is located at a specific preferred distance and allocentric direction. The firing rate of a BVC neuron $i$, tuned to a preferred distance $d_i$ and direction $\phi_i$, is given by:

\vspace{-2mm}
\begin{equation}
v_i^b = \sum_{j=0}^{n_\text{res}}
\left(
\frac{\exp \left[-\frac{(r_j - d_i)^2}{2 \sigma_r^2} \right]}{\sqrt{2 \pi \sigma_r^2}}
\times
\frac{\exp \left[-\frac{(\theta_j - \phi_i)^2}{2 \sigma_\theta^2} \right]}{\sqrt{2 \pi \sigma_\theta^2}}
\right),
\end{equation}

\noindent
where $r_j$ and $\theta_j$ represent the distance and direction of an environmental boundary point $j$, while $\sigma_r$ and $\sigma_\theta$ parameterize the tuning widths for the distance and angular components, respectively. The summation over $j$ aggregates responses across all boundary points within the environment, with $n_\text{res}$ representing the resolution of boundary encoding. To support true 3D representations, a vertical sensitivity component was added to allow BVCs to incorporate not only horizontal angles but also vertical pitch information. This additional dimension helps distinguish overlapping 2D features when ceilings, sloped walls, or multi-level structures are present. The extended firing rate equation is:

\vspace{-2mm}
\begin{align}
v_i^b &= \frac{1}{\text{norm}} \sum_{j=0}^{n_\text{res}}
\Bigg(
\frac{\exp \left[-\frac{(r_j - d_i)^2}{2 \sigma_r^2} \right]}{\sqrt{2 \pi \sigma_r^2}}
\times \nonumber \\
& \qquad
\frac{\exp \left[-\frac{(\theta_j - \phi_i)^2}{2 \sigma_\theta^2} \right]}{\sqrt{2 \pi \sigma_\theta^2}}
\times
\frac{\exp \left[-\frac{(\phi_j - \psi_i)^2}{2 \sigma_\phi^2} \right]}{\sqrt{2 \pi \sigma_\phi^2}}
\Bigg),
\end{align}

\noindent
where $\phi_j$ represents the vertical angle of a boundary point, $\psi_i$ is the preferred vertical angle for the BVC neuron, and $\sigma_\phi$ is the tuning width for the vertical sensitivity. The normalization factor, $\text{norm}$, scales the activations to ensure that they fall within the range $[0, 0.5]$

BVC cells' preferred distances are determined by their placement along an axis of their preferred directions, spaced evenly to a maximum distance with a specified number of BVCs per direction. The number of cells to place along an axis is a hyperparameter of the model.

The inclusion of vertical sensitivity enhances the model's ability to disambiguate overlapping features in environments with significant vertical structure by introducing the ability to add additional layers of BVCs arranged vertically.

\subsection{Place Cell Network}

The cells of the PCN forms place fields in a given environment through exploration. Each place cell combines weighted excitatory input from several BVCs, subject to both feedforward inhibition from the total BVC activity and recurrent inhibition from the place cell network itself. The dynamics of the membrane potential $s_i^p$ for the $i$-th place cell are described by:

\vspace{-1mm}
\begin{align}
    \tau_p \frac{ds_i^p}{dt} &= -s_i^p 
    + \sum_{j=0}^{n_b} W_{ij}^{pb} v_j^b \nonumber \\
    &\quad - \Gamma_{pb} \sum_{j=0}^{n_b} v_j^b 
    - \Gamma_{pp} \sum_{j=0}^{n_p} v_j^p,
\end{align}

where:
\begin{itemize}
    \item $\tau_p$ is the time constant for place cell membrane potential dynamics
    \item $s_i^p$ is the membrane potential of the $i$-th place cell
    \item $W_{ij}^{pb}$ is the synaptic weight from the $j$-th BVC to the $i$-th place cell
    \item $v_j^b$ is the firing rate of the $j$-th BVC
    \item $\Gamma^{pb}$ is the gain parameter for feedforward inhibition from BVCs
    \item $\Gamma^{pp}$ is the gain parameter for recurrent inhibition from other place cells
    \item $v_j^p$ is the firing rate of the $j$-th place cell
\end{itemize}

The firing rate $v_i^p$ of the $i$-th place cell is given by a rectified hyperbolic tangent of its membrane potential, scaled by a gain parameter $\psi$:

\vspace{-1mm}
\begin{equation}
v_i^p = \tanh \left( \psi [s_i^p]_+ \right),
\label{eq:place_cell_firing_rate}
\end{equation}

where $[s_i^p]_+ = \max(s_i^p, 0)$ is the rectified membrane potentials.

Feedforward inhibition, represented by the term $\Gamma^{pb} \sum_{j=0}^{n_b} v_j^b$, sets a threshold on the amount of input required for a place cell to activate, while the recurrent inhibition term $\Gamma^{pp} \sum_{j=0}^{n_p} v_j^p$ prevents uncontrolled network activity by suppressing overactive place cells. Together, these inhibitory mechanisms ensure spatial specificity and stability in the place cell network.

To achieve uniform coverage over the environment, competitive learning is implemented using the rule proposed by Oja \cite{oja1982simplified}. The synaptic weight $W_{ij}^{pb}$ from the $j$-th BVC to the $i$-th place cell evolves as:

\vspace{-2mm}
\begin{equation}
    \tau_{w^{pb}} \frac{dW_{ij}^{pb}}{dt} = v_i^p \left(v_j^b - \frac{1}{\alpha_{pb}} v_i^p W_{ij}^{pb}\right),
\end{equation}

where:
\begin{itemize}
    \item $\tau^{wpb}$ is the time constant for learning dynamics.
    \item $\alpha^{pb}$ is a normalizing factor for synaptic weights.
    \item $v_i^p$ is the firing rate of the $i$-th place cell.
    \item $v_j^b$ is the firing rate of the $j$-th BVC.
\end{itemize}

In this competitive learning framework, synapses are potentiated based on the product of BVC and place cell firing rates, promoting stronger connections to active BVCs. A forgetting term proportional to the square of the place cell output ensures that synaptic weights remain bounded, preventing divergence and promoting competition. This mechanism allows inactive place cells to ``win'' the competition when their corresponding BVCs become active, resulting in the formation of localized place fields.


\section{Methods}

\subsection{Simulation Setup}

This experiment was conducted in simulation on the Webots platform version 2021a \cite{Webots}. The environment was a square arena, $10 \times 10$ meters in size, enclosed by vertical walls, each 2.5 meters tall, and a ceiling. For visualization purposes, the ceiling was rendered transparent, but it remained detectable by the agent's distance sensors. The environment had two central walls forming a cross shape, each extending 7 meters in length. In all scenarios, the walls extended to the ceiling.

\begin{figure}[h!]
    \centering
    \begin{subfigure}[b]{0.45\linewidth}
        \centering
        \includegraphics[width=\linewidth]{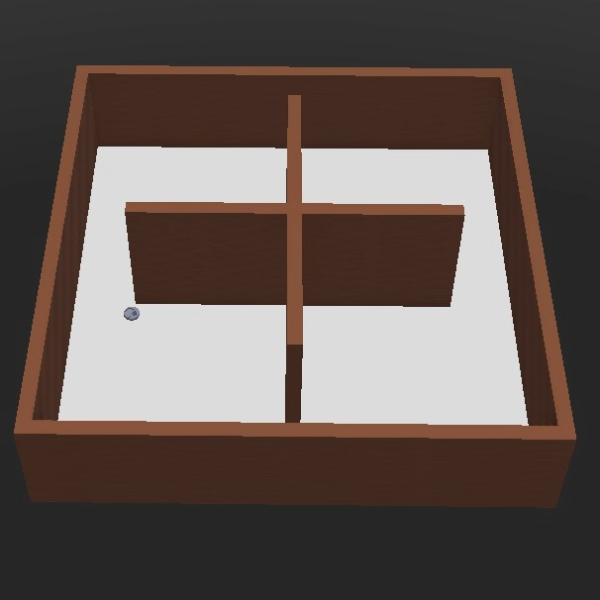}
        \caption{Environment 1: Baseline Upright walls}
        \label{fig:env1}
    \end{subfigure} \quad
    \begin{subfigure}[b]{0.45\linewidth}
        \centering
        \includegraphics[width=\linewidth]{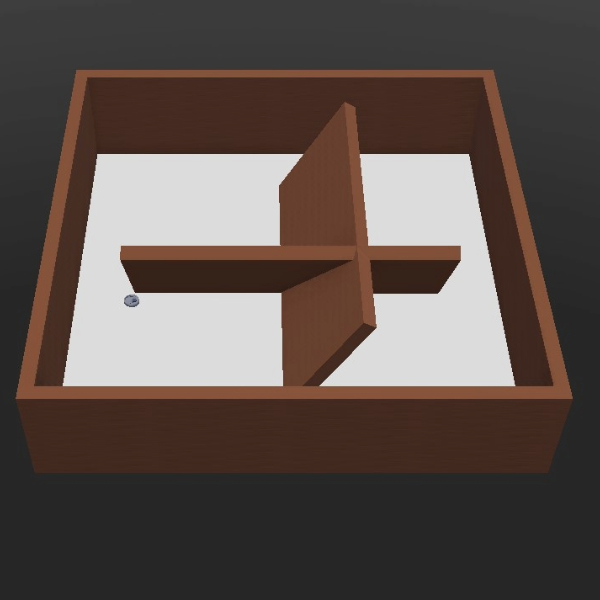}
        \caption{Environment 2: Tilted 30° Mild variation}
        \label{fig:env2}
    \end{subfigure}

    \begin{subfigure}[b]{0.45\linewidth}
        \centering
        \includegraphics[width=\linewidth]{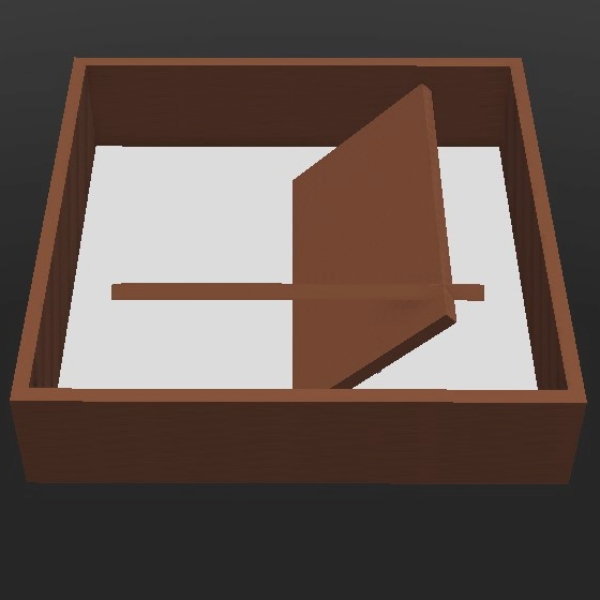}
        \caption{Environment 3: Tilted 45° Moderate variation}
        \label{fig:env3}
    \end{subfigure} \quad
    \begin{subfigure}[b]{0.45\linewidth}
        \centering
        \includegraphics[width=\linewidth]{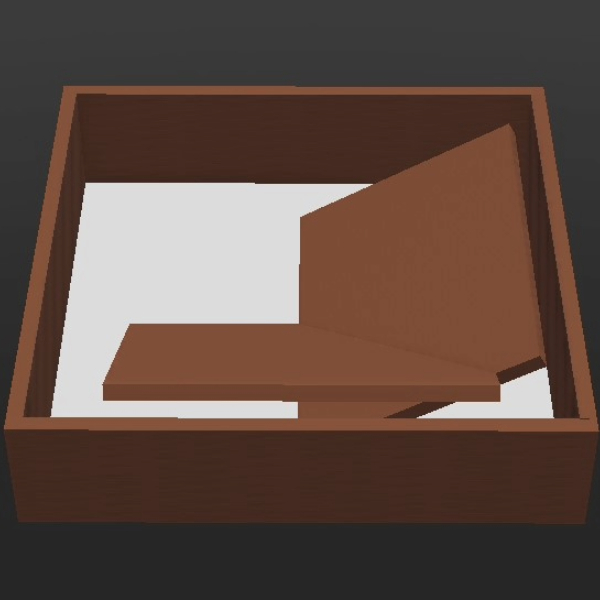}
        \caption{Environment 4: Tilted 60° Extreme variation}
        \label{fig:env4}
    \end{subfigure}

    \caption{Illustrations of the four test environments with varying wall orientations, designed to evaluate the model's 3D spatial encoding capabilities. The ceiling is transparent in the visualizations to allow a clearer view of the environments.}
    \label{fig:environments}
\end{figure}

To increase the 3D complexity of the environment while preserving the floor area and maintaining the 2D layout's consistency, the central walls were both rotated in equivalent degree intervals towards one corner. The test environments were as follows (Figure~\ref{fig:environments}):

\begin{itemize}
    \item \textbf{Environment 1:} Baseline configuration with upright walls.
    \item \textbf{Environment 2:} Central walls tilted 30°, introducing mild vertical variation.
    \item \textbf{Environment 3:} Central walls tilted 45°, introducing moderate vertical variation.
    \item \textbf{Environment 4:} Central walls tilted 60°, introducing extreme vertical variation.
\end{itemize}

\vspace{-2mm}

\subsection{Agent Description}

The agent was modeled after the Create series of Roomba robots, with a cylindrical body approximately 0.5 meters in diameter. It is capable of forward, backward, and zero-axis turn rotation. Two distance scanners are available on the robot. The first scanner is a horizontal distance scanner that captures a set of 720 scan points in 360 degrees along the horizontal plane. 
Additionally, a second spherical distance scanner, capable of generating a depth map image of $90 \times 180$ pixels, is mounted on the agent. This scanner captures distances in every direction horizontally and vertically. 
To prevent false detections of nearby obstacles or detecting the floor as an obstacle by the BVCs, the bottom half of the data was discarded. This ensured that any BVCs did not incorrectly detect the agent's body or the floor at every time step. 

The agent was programmed to explore throughout the environment in a random walk, which was performed as follows: 

The agent moved forward for a predefined number of steps, $\tau_w$, while updating its place
cell activations. If a collision was detected via bumper sensors, the agent selected a new heading angle uniformly at random between $[-\pi, \pi]$ radians, turned to that direction, and resumed forward motion. Additionally, after completing $\tau_w$ time steps without collision, the agent chose a new direction by sampling an angle from a Gaussian distribution centered at $0$ with a standard deviation of $30^{\circ}$ (or $\pi/6$ radians). This strategy enabled the agent to explore the environment thoroughly, with periodic changes in direction to avoid stagnation.

\begin{figure}[h!]
    \centering
    \includegraphics[width=50mm]{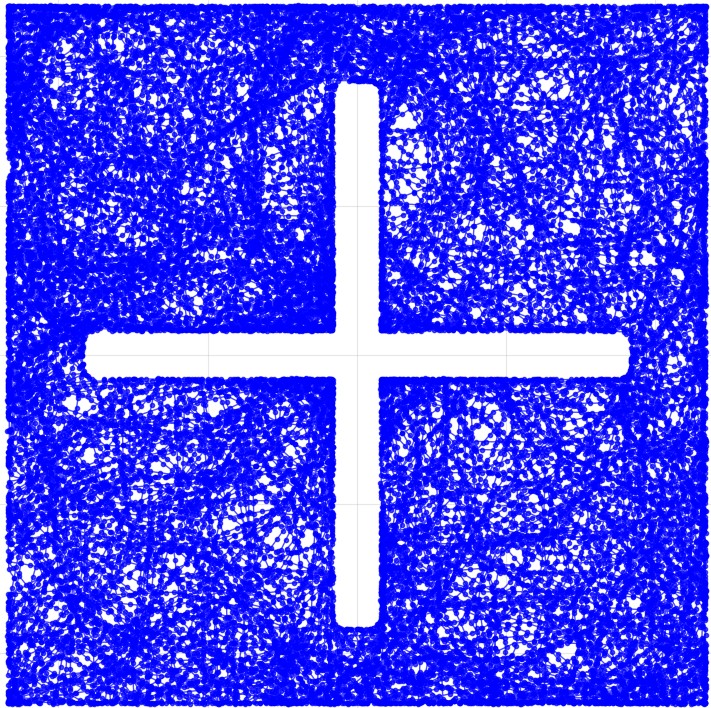}
    \caption{A sampling trajectory plot of the agent's random walk over 4 hours, demonstrating dense and uniform coverage across the environment.}
    \label{fig:trajectory}
\end{figure}

To validate the learning that occurred in the exploration phase, the agent entered a 4-hour random walk {\it sampling session} to obtain thorough coverage of the entire environment (which is not guaranteed during the exploration phase), ensuring every point in the environment was visited at least once. A sample trajectory plot from this run, shown in Figure~\ref{fig:trajectory}, illustrates the efficiency of the random walk algorithm in ensuring comprehensive coverage.

\vspace{-3mm}

\subsection{Model Configurations}
\subsubsection{Parameter Selection}  
To ensure consistency in the number of parameters across all models, each configuration was initialized with 960 BVCs and 250 place cells.
The BVCs were arranged along a maximum distance threshold of 12 meters with a tuning parameter $\sigma_d$ set to 0.75 meters to maintain similarly sized place fields.
The total number of BVCs was kept constant by adjusting the number of cells allocated to each preferred directional axis.
The angular tuning parameter \(\sigma_\theta = 0.1\)~radians
in both 2D and 3D models.
For the vertical sensitivity, the tuning parameter $\sigma_\phi$ was set to 0.01 radians to capture fine-grained vertical variations. 

The vertical angular tuning width \(\sigma_\phi = 0.01\)~radians was chosen based on preliminary experiments showing that larger values (e.g., \(\sigma_\phi > 0.05\)) led to overlapping vertical detections, degrading place field specificity. This tight tuning aligns with commercial LiDAR architectures such as the Velodyne HDL-64E, which samples vertical angles at fixed resolutions (e.g., \(0.1\)--\(0.4\)~radians per layer).  

Horizontal angular tuning  was reduced from earlier work (\(\sigma_\theta = 1.57\)~radians) to mitigate aliasing caused by non-proximal wall segments. Wider horizontal tuning (\(\sigma_\theta > 0.2\)) increased false boundary associations, as BVCs responded to distant LiDAR points misaligned with actual walls. This sensitivity arises because broader angular receptive fields integrate more boundary points, amplifying noise from sparse or oblique detections.

The specific model configurations were as follows:

\renewcommand{\arraystretch}{1.0}
\setlength{\tabcolsep}{4pt}

\begin{table}[h!]
\caption{Model Configurations}
\label{tab:models}
\centering
\begin{tabular}{|l|c|c|c|c|}
\hline
\textbf{Model} & \makecell{\textbf{Horizontal} \\ \textbf{Directions}} & \makecell{\textbf{Vertical} \\ \textbf{Directions}} & \makecell{\textbf{Vertical} \\ \textbf{Angles (rad)}} & \makecell{\textbf{BVCs} \\ \textbf{per Axis}} \\ \hline
2D & 8 & 0 & 0.0 & 120 \\ \hline
3D (0.1 rad) & 8 & 2 & 0.0, 0.1 & 60 \\ \hline
3D (0.2 rad) & 8 & 2 & 0.0, 0.2 & 60 \\ \hline
3D (three-layer) & 8 & 3 & 0.0, 0.1, 0.2 & 40 \\ \hline
\end{tabular}
\end{table}

These configurations ensured consistent parametrization while testing the impact of increasing vertical sensitivity on spatial encoding.
By systematically varying the number and orientation of vertical layers, we evaluated the role of vertical sensitivity in enhancing the model's spatial representation and ability to disambiguate complex environments.
The decision for the elevations was made 

\subsubsection{Vertical Layer Configuration}  
Vertical layer orientations (\(0.1\) and \(0.2\)~radians above the horizontal plane) were systematically selected to test incremental gains from added 3D sensitivity rather than through optimization. By comparing models with one or two added layers against the 2D baseline, we isolated the impact of vertical complexity on spatial encoding without presuming ideal layer heights. 

\subsection{Data Collection}
Each of the four models was evaluated in each of the four environment
configurations. This yielded a total of 16 experimental trials. 
Each trial was run for a simulation time of 4 hours. During each trial, the agent’s \((x,y)\) position and the place cell activations were recorded when the agent's BVC and PCN were updated, producing, on average, around 30 thousand measurements per trial. 
These points were then preprocessed and analyzed as outlined in the next two sections.

\vspace{-0.25mm}
\subsection{Data Processing and Analysis}

\subsubsection{Hexmap Creation}
The environment was discretized into \(50 \times 50\) hexagonal bins. Each place cell's activation in each bin was computed as the mean firing rate across visits to thatbin, thresholded at the 10\% quantile to remove noise, and normalized to \([0, 1]\).

In the resulting activation map, each bin’s center \((x_i, y_i)\) was associated with the corresponding mean activation value \(a_i\) for that bin.

\vspace{-0.25mm}
\subsection{Analysis Metrics}

We used two principal metrics to evaluate how effectively each model’s place cells localized the robot. Ideally, a place cell should have a well-defined, roughly circular unimodal (single-peaked) firing field anchored to a single region in the environment. The two metrics are:

\begin{itemize}
    \item \textbf{Modality Index}: This assesses whether each place cell’s firing distribution is unimodal or multimodal, using a clustering algorithm.
    \item \textbf{Spatial Aliasing Index}: This metric quantifies the similarity of spatially distant locations' place-cell activation patterns at the bin level. The metric was then averaged to measure performance over the whole environment.
\end{itemize}

\subsubsection{Modality Index}
The Modality Index (MI) for a given place cell was calculated by applying the DBSCAN clustering algorithm to the cell's activation map to identify significant clusters of activity. The radius for DBSCAN clustering was set to $\epsilon = 1$, and the minimum number of samples required within that radius to $20$ to ensure robust clustering. Each identified cluster represented a mode in the activation map of the place cell, and the modality index, $\mathrm{MI} (m)$ for cell $m$ was set to the number of clusters found. For a place cell with unimodal firing, \(\mathrm{MI} (m) = 1\).

The overall model performance using the MI was measured using three metrics derived from the cell modality indices:

\begin{itemize}
    \item \textbf{Fraction of Cells with \(\mathrm{MI} > 0\)}:
    The fraction of all place cells that have a nonzero firing field (i.e., at least one detected cluster).
    \item \textbf{Average \(\mathrm{MI}\) of Nonzero Cells}:
    The modality index calculated only over cells with \(\mathrm{MI} > 0\).
    \item \textbf{Fraction of Cells with \(\mathrm{MI} > 1\)}:
    The fraction of all place cells that exhibit more than one cluster.
\end{itemize}
Lower values of the latter two metrics (i.e., fewer overall clusters and fewer cells with multiple fields) imply more stable single-field place cells and thus better spatial specificity.

\subsubsection{Spatial Aliasing Index and Mean Spatial Aliasing Index}
The Spatial Aliasing Index (SAI) and Mean Aliasing Index (MSAI) capture how the entire place cell ensemble responds across different regions. These metrics quantify the similarity between spatially distant bins' activation patterns.

\paragraph{SAI}
For a bin \(i\) located at \((x_i, y_i)\), the SAI quantified the degree of spatial aliasing in place representations across the environment. It is given by:

\vspace{-2mm}
\begin{multline}
    \mathrm{SAI}^{(i)} = \frac{1}{N} \sum_{j=1, j \neq i}^{N} \mathbf{1}\bigl[\|(x_i, y_i) - (x_j, y_j)\| > d_{\mathrm{th}}\bigr] \\
    \cdot \mathrm{CosSim}\bigl(\mathbf{a}^{(i)}, \mathbf{a}^{(j)}\bigr),
\end{multline}
\vspace{-1mm}
where:
\begin{itemize}
    \item \(N\) is the total number of bins.
    \item \(d_{\mathrm{th}}\) is the distance threshold to exclude nearby bins.
    \item \(\mathbf{1}[\cdot]\) is the indicator function (1 if the distance between bins exceeds \(d_{\mathrm{th}}\), 0 otherwise).
    \item \(\mathrm{CosSim}(\mathbf{a}^{(i)}, \mathbf{a}^{(j)})\) is the cosine similarity between the place cell activation vectors in bins \(i\) and \(j\).
\end{itemize}
A higher \(\mathrm{SAI}^{(i)}\) indicated that the bin’s place cell activation pattern is more similar to those in distant regions, suggesting potential aliasing. A lower \(\mathrm{SAI}^{(i)}\) indicated more distinct activations, reflecting better spatial specificity.

\paragraph{MSAI}
To evaluate the model’s overall aliasing performance, we computed the MSAI as the average of the SAI across all bins:
\vspace{-3mm}
\begin{equation}
    \mathrm{MSAI} = \frac{1}{N} \sum_{i=1}^{N} \mathrm{SAI}^{(i)}
\end{equation}
A higher \(\mathrm{SAI}^{(i)}\) or \(\mathrm{MSAI}\) suggests more aliasing, indicating that distant bins exhibited similar activation patterns and weaker localization, while lower values of \(\mathrm{SAI}^{(i)}\) or \(\mathrm{MSAI}\) implied better spatial discrimination and stronger place-cell localization.


\section{Results}

Results This section reports results on the performance of the four different models across four increasingly complex 3D environments. While we present results for one representative set of runs in each environment to compare the models, other tests have shown consistent qualitative trends across multiple trials, reinforcing the robustness of the findings. 

\subsection{Modality Index (MI) Results}

\begin{figure}[H]
\centering
\includegraphics[width=80mm]{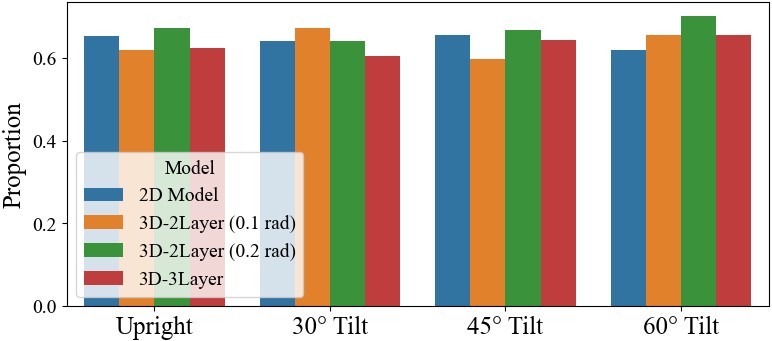}
\caption{Percentage of place cells with \(\mathrm{MI} > 0\). Across all models and environments, the proportion of active cells remains consistently high.}
\label{fig:mi_percent}
\end{figure}
\begin{figure}[H]
\centering
\includegraphics[width=80mm]{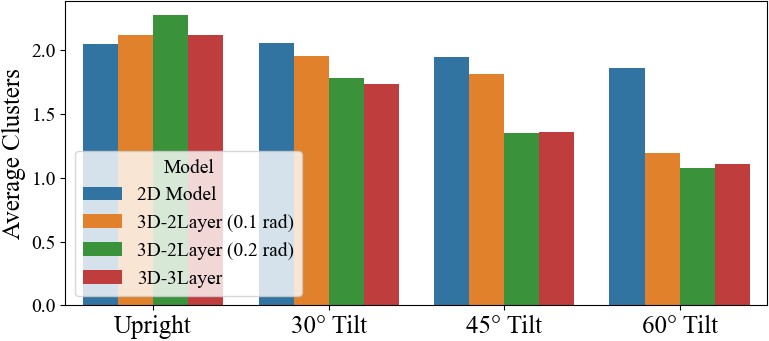}
\caption{Average MI across models and environments. The 2D model maintains a mean MI near 2.0 (two modes per cell), whereas the 3D models exhibit a marked reduction in MI as the wall tilt increases, reaching below 1.25 by the 60\degree{}.}
\label{fig:mi_average}
\end{figure}
\begin{figure}[H]
\centering
\includegraphics[width=80mm]{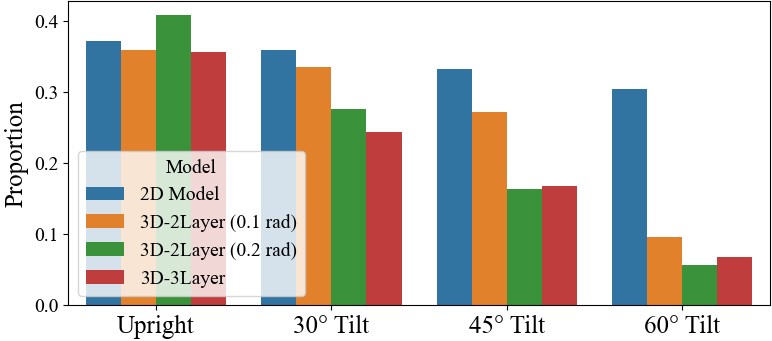}
\caption{Fraction of place cells with \(\mathrm{MI} > 1\), indicating multi-modal cells. The 2D model consistently has around 30--40\% multi-modal cells, whereas the 3D models drop to about 5\% by the 60\degree{} tilt.}
\label{fig:mi_multiple}
\end{figure}

Figures~\ref{fig:mi_percent}--\ref{fig:mi_multiple} show the results using the modality index metrics.

\vspace{-4mm}

\vspace{-4mm}

\vspace{-4mm}

\paragraph{Key Observations:}
\begin{itemize}
    \item \(\mathrm{MI} > 0\) remains stable across all models and environments, indicating that each model activates a similar proportion of place cells. This indicates that differences in the other metrics describing representational quality are not due to changes in the active population of cells.
    \item The 3D models significantly reduce the number of multi-modal cells (\(\mathrm{MI} > 1\)) and lower the average MI compared to the 2D baseline, demonstrating improved formation of distinct (unimodal) firing fields.
    \item When the environment’s vertical component is small (e.g., upright walls), all models show similar MI values. As the tilt increases, the 3D models outperform the 2D model, highlighting the advantage of vertical sensitivity in more complex 3D scenes.
\end{itemize}

\subsection{Aliasing Index (SAI) Results}

\begin{figure}[H]
\centering
\includegraphics[width=88mm]{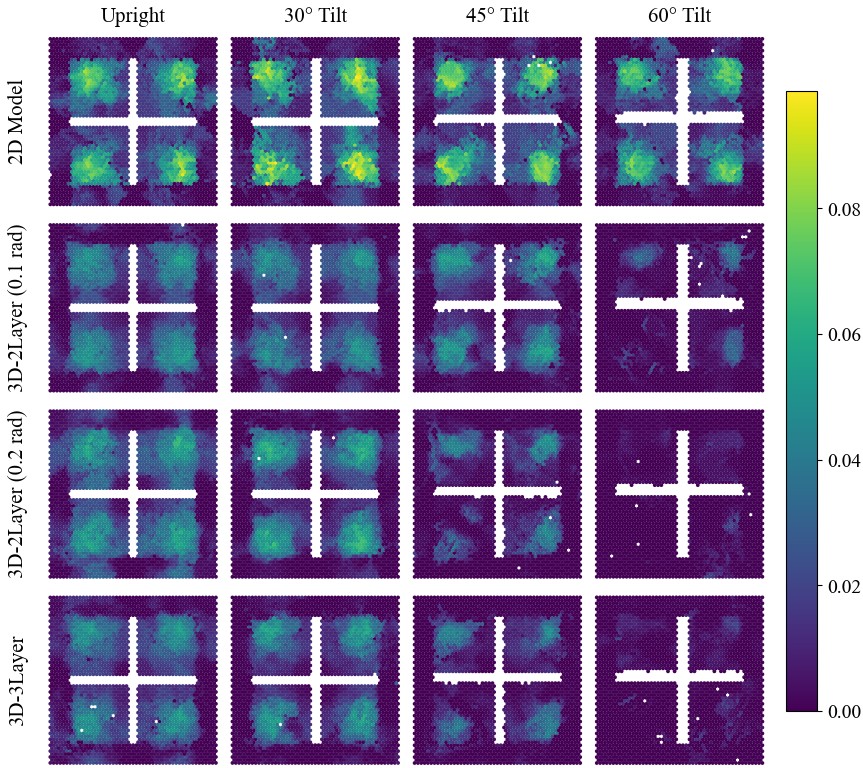}
\caption{Grid of 16 subplots showing the SAI values across all bins for each model and environment. Brighter regions indicate locations with higher aliasing. The 2D model shows more extensive and constant aliasing across environments due to the symmetric structure of the environment, whereas the 3D models display progressively lower aliasing, especially the 3-layer and 0.2-radian models.}
\label{fig:ai_subplots}
\end{figure}

Next, we visualize the SAI values for each bin in the environment as spatial heatmaps (Fig.~\ref{fig:ai_subplots}). A higher SAI means more similarity with distant bins, reflecting increased aliasing.

\paragraph{Key Observations}
\begin{itemize}
    \item The 2D model shows distinct bright zones (high aliasing) across all environments.
    \item The 3D two-layer (0.2 rad) and three-layer models exhibit the fewest bright regions, suggesting improved spatial discrimination due to additional vertical sensitivity.
    \item As the wall tilt becomes more pronounced, the benefit of having vertical layers is more apparent in reducing overlapping place cell responses across distant locations.
\end{itemize}

\subsection{MSAI Results}

The MSAI aggregates the overall aliasing of all locations in an environment, with lower values denoting improved spatial discrimination.

\begin{figure}[ht]
\centering
\includegraphics[width=80mm]{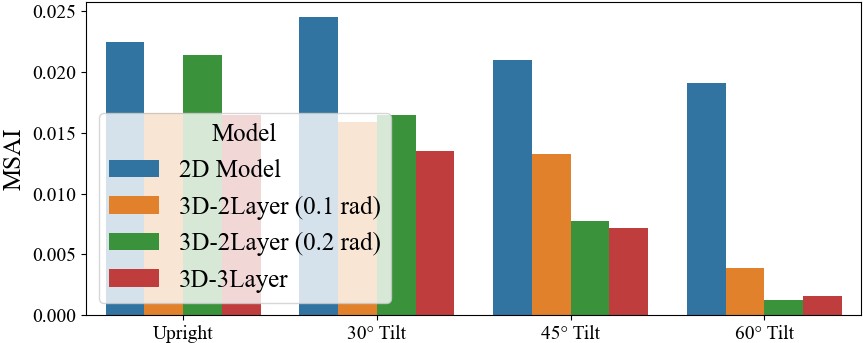}
\caption{MSAI for each model in each environment. The 3D models maintain consistently lower MSAI values, indicating better overall spatial discrimination compared to the 2D baseline.}
\label{fig:mai_bar}
\end{figure}

\paragraph{Key Observations:}
\begin{itemize}
    \item While the 2D model shows only slight MSAI reductions across environments, the 3D models achieve significantly lower MSAI values, especially at higher wall tilts.
    \item The three-layer and two-layer (0.2 rad) configurations generally yield the lowest MSAI, suggesting that additional vertical sensitivity translates to more robust localization.
\end{itemize}



In summary, the results suggest that adding even moderate vertical sensitivity to the BVC model leads to more distinct place fields and lower aliasing in complex 3D scenarios. In environments where vertical dimensions are less relevant, the 3D-enhanced models exhibit performance similar to the 2D baseline, confirming that the added vertical parameter does not degrade performance in simpler cases.


\section{Discussion}
In this work, we introduced and evaluated a three-dimensional extension of the BVC model by adding BVC layers that are vertically oriented at specific angles relative to the horizontal plane. Our results show that incorporating these additional vertical orientations can substantially enhance spatial encoding in environments with pronounced 3D structure. The key findings are as follows:

\begin{itemize}
    \item \textbf{Superiority of Steeper Vertical Orientations:} Among the 3D configurations, the models featuring a set of BVCs oriented at 0.2\,rad above the horizontal consistently produced the most robust place cell activity, with nearly all cells exhibiting unimodal activity. This suggests that a wider sensing angle in the vertical direction enhances performance, though the exact relationship between angular field width and performance likely depends on the specific 3D geometry of the environment.

    \item \textbf{Marginal 2D Gains:} Although the 2D model showed minimal improvements as the central walls were rotated, it also generally failed to leverage any additional vertical cues. As the tilt angle increased, the 3D models—with their vertical BVC orientations—became markedly more effective at disambiguating different parts of the environment.

    \item \textbf{Impact of Subtle 3D Complexity.} Even mild tilts in the central walls introduced enough vertical complexity to significantly challenge the 2D baseline. The presence of vertically oriented BVC layers helped distinguish overlapping 2D features when the walls were rotated, thereby preventing spatial aliasing that would otherwise occur in the 2D model.

\end{itemize}

Despite these promising outcomes, there are still open questions about the model’s generalizability to real-world sensor noise, negative values, complex 3D structures, and changing boundary heights. Additionally, exploring multi-layer BVC orientations beyond 0.1\,rad and 0.2\,rad, such as finer-grained increments, could further refine how we capture vertical structure. Finally, integration with more biologically inspired components, such as a vision system or a grid cell network, would help to develop a more comprehensive 3D spatial navigation framework.


\section{Conclusions and Future Work}


Overall, our findings demonstrate that adding select vertical orientations to BVCs can substantially enhance an agent’s capacity for spatial localization in complex 3D settings. Future investigations will explore optimal strategies for determining these orientations, integrating adaptive mechanisms, and extending the approach to more varied real-world conditions and multimodal sensor inputs. 
In particular, it would be valuable to investigate how the degree of 3D sensitivity should depend on the geometric characteristics of the environment, potentially leading to an adaptive or learning-based approach for selecting vertical orientations. 
Additionally, it would be interesting to determine if it is possible to orient layers of BVCs downwards to create 3D place cell formations utilizing flying agents.
The future investigations will take the model closer to a comprehensive 3D spatial navigation model that mirrors the complexity and flexibility observed in biological systems.


\section*{Acknowledgments}

The authors gratefully acknowledge Adedapo Alabi for several very helpful discussions and explanations of the previous model which was used as the basis for the more complex model described in this paper.

The conceptualization, methodology, experimental design, simulations, data analysis, interpretation of results, and the primary textual content in this paper are the work of the authors. AI tools, specifically OpenAI’s ChatGPT-4o and o1, were used to provide assistance limited to formatting figures, equations, and tables, grammar checking, and minor corrections.

\balance


\end{document}